\documentclass[letterpaper, 10 pt, conference]{ieeeconf}  

\IEEEoverridecommandlockouts                              
\usepackage{graphicx} 

\usepackage{array}    
\usepackage{siunitx}
\usepackage{subfig}
\usepackage{soul,xcolor}
\usepackage{amssymb}
\usepackage{amsmath} 
\usepackage{pifont}
\usepackage{float}

\usepackage{algorithm}
\usepackage[noend]{algpseudocode}
\usepackage{mathtools}
\usepackage{amsmath} 
\usepackage{color}
\usepackage{url}

\newcommand{\IB}[1]{\textcolor{black}{#1}}
\setstcolor{red}

\title{\LARGE \bf
Context-Dependent Anomaly Detection for Low Altitude Traffic Surveillance 
}

\author{Ilker Bozcan and Erdal Kayacan

\thanks{This work was financially supported by Aarhus University, Department of  Electrical and Computer Engineering (28173).}
\thanks{I. Bozcan and E. Kayacan are with the Artificial Intelligence in Robotics Laboratory (Air Lab), Department of Electrical and Computer Engineering, Aarhus University,
        8000 Aarhus C, Denmark
        {\tt\small \{ilker, erdal\} at ece.au.dk}}%
}

\begin{document}

\maketitle
\thispagestyle{empty}
\pagestyle{empty}

\begin{abstract}

\IB{The} detection of contextual anomalies is a challenging task for surveillance since an observation can be considered anomalous or normal \IB{in} a specific environmental context. \IB{An unmanned aerial vehicle (UAV) can utilize its aerial monitoring capability and employ multiple sensors to gather contextual information about the environment and perform contextual anomaly detection.} In this work, we introduce a deep neural network-based method (CADNet) to find point anomalies (i.e., single instance anomalous data) and contextual anomalies (i.e., context-specific abnormality) in an environment using a UAV. The method is based on a variational autoencoder (VAE) with a context sub-network. The context sub-network extracts contextual information regarding the environment using GPS and time data, then feeds it to the VAE to predict anomalies conditioned on the context. To the best of our knowledge, our method is the first contextual anomaly detection method for UAV-assisted aerial surveillance. We evaluate our method on the AU-AIR dataset \IB{in} a traffic surveillance scenario. Quantitative comparisons against several baselines demonstrate the superiority of our approach in the anomaly detection tasks. The codes and data will be available at \url{https://bozcani.github.io/cadnet}.







\end{abstract}

\section{INTRODUCTION}



Anomaly detection is defined as the task of finding abnormal patterns in data that are significantly different from the vast majority of observations. Although the definition of the task is explicit, it is not very straightforward to determine a clear description of anomalies. However, regardless of their descriptions, anomalies can be categorized into three types: point anomalies (i.e., single data instance as an anomaly), collective anomalies (i.e., collection of data instances as an anomaly), and contextual anomalies \cite{chandola2009anomaly}. Contextual anomalies are data instances that are considered \IB{anomalous} conditioned on a specific context. Unlike point and collective anomalies, contextual anomalies may be considered normal in a different context. This phenomenon makes the contextual anomaly detection task more difficult compared to point or sequential anomaly detection. In this work, we investigate the contextual anomaly detection task for a  surveillance scenario.



In the context of surveillance, observations of instances (e.g., people, vehicles), \IB{which break} the public rules (e.g., traffic rules), in single image or frame sequences can be considered  point and sequential anomalies, respectively. These types of anomalies are straightforward since the precise definitions of anomalies are available. However, the determination of contextual anomalies can be entirely formidable due to the lack of clarity. Contextual anomalies in an environment do not necessarily break any rules, whereas they arise suspicions under certain conditions. For example, a car may be considered normal in a car parking area \IB{yet} it is an anomaly in case of an appearance in an abandoned land.
\begin{figure}[!t] 
\centerline{
\includegraphics[width=0.49\textwidth]{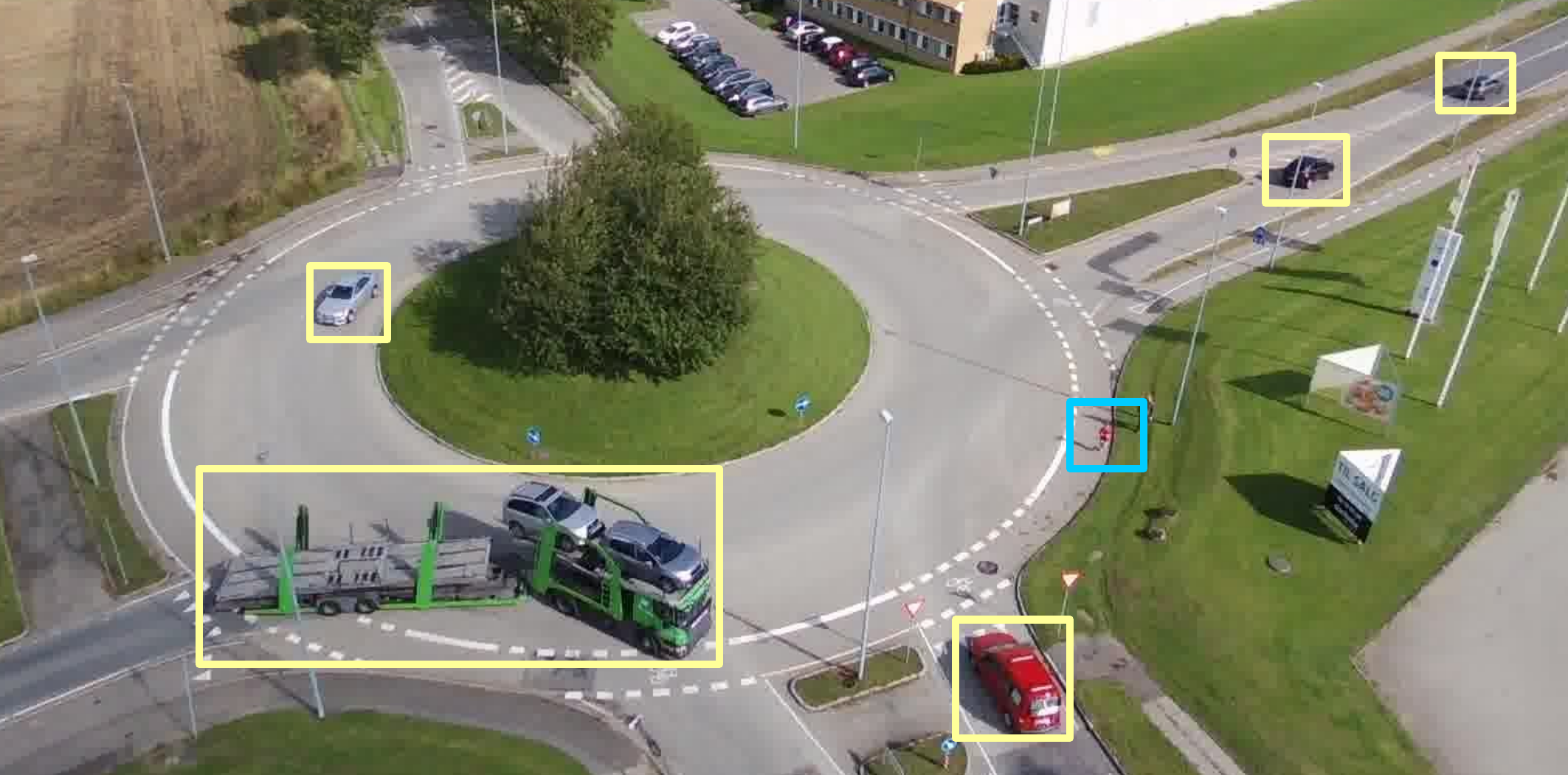}
}
\caption{An example aerial image captured by a UAV. The monitoring capability with a large view makes a UAV significantly useful for surveillance. For a traffic surveillance case, they can monitor a large number of vehicles \IB{(yellow boxes)} and pedestrians \IB{(blue box)}. The image is a sample from the AU-AIR dataset \cite{bozcan2020air}. \label{fig:intro}}
\end{figure}

Naturally, contextual anomaly detection is more challenging than a point or collective anomaly detection task. Firstly, contextual attributes should contain relevant contextual information regarding an environment, and be appropriately chosen according to the application domain. For instance, the longitude and latitude can give a significant clue about the environmental context in case of anomaly detection task on spatial data for surveillance. Then, contextual information should be adequately fused into an anomaly detector. Hence, context-awareness is essential for anomaly detection systems for surveillance applications.

Unmanned aerial vehicles (UAVs) equipped with a camera, a GPS sensor, and an on-board computer can be deployed to a surveillance system to conduct context-aware autonomous anomaly detection tasks (See Fig. \ref{fig:intro}). The internal clock of the on-board computer and GPS sensor can supply contextual information regarding time and spatial location for an environment. As a result of flight capability, a UAV overcomes significant limitations of current video-based surveillance systems, which are based on fix cameras (e.g., lack of taking a closer look, anomaly tracking, and limited coverage).

In this work, we present a deep neural network-based context-aware anomaly detection method (CADNet) for aerial video surveillance with a UAV. The method works at the top of any YOLO-variant object detector (e.g., YOLO \cite{redmon2016you}, YOLOv3 \cite{redmon2018yolov3}, Tiny-YOLOv3 \cite{redmon2018yolov3}) whose final output is a fixed-size tensor. CADNet takes the tensor slice, which includes class predictions from the object detector's output, as the main input (See Fig. \ref{fig:architecture}). After the inference, it removes predictions, which correspond to anomalies in the image, from the tensor. Moreover, we consider time and GPS location, which are related to aerial surveillance scenarios, as contextual attributes. These attributes give a clue to the network regarding the environmental context in order to find context-dependent anomalies. Therefore, we employ a context subnetwork to extract features from raw time and GPS data. 

\subsection{Related Work}


Most of the studies conduct anomaly detection tasks using a stationary surveillance camera \cite{hasan2016learning, sultani2018real,xu2015learning,gao2016violence,kooij2016multi}. Stationary surveillance cameras have a problem of immobility that prevents the detection of location-dependent contexts. There are also studies using a pan-tilt camera \cite{benito2018deep}, that lacks mobility, and moving cameras \cite{de2019anomaly, nakahata2018anomaly} mounted on inspection robots. Henrio et al. \cite{henrio2018anomaly} and Singh et al. \cite{henrio2018anomaly} use a UAV that can conduct a wide range of monitoring. 

Some of the studies calculate an anomaly score for a given input, using an autoencoder (AE) \cite{hasan2016learning}, multiple instance learning (MIL) \cite{sultani2018real}, stack denoising autoencoders (SDAE) \cite{xu2015learning}, one-class support vector machine (SVM) with feature extraction (FE) \cite{gao2016violence}, convolutional neural networks (CNNs) and recurrent neural networks (RNNs) \cite{henrio2018anomaly}. However, instead of finding individual anomalous entities at a semantic level, they infer anomalies at a lower level (e.g., pixel-level anomalies). Therefore, they lack an explicit representation of anomalies. Nakahata et al. \cite{nakahata2018anomaly} propose a spatio-temporal composition (STC) method to detect anomalous pixels in video frames instead of calculating an anomaly score. Unlike other methods, Kooij et al. \cite{kooij2016multi} \IB{use} Bayesian Networks (BN) to combine audial and visual features for anomaly detection. Singh et al. \cite{singh2018eye} use ScatterNet hybrid deep learning (SHDL) for human pose estimation and then apply SVM to identify violent behaviors. However, they solely use visual data as input. Bozcan et al. \cite{bozcanilkeruavadnet} propose a variant of a variational autoencoder to find anomalies in a meta-representation of images. Unlike others, they find anomalies in object-level. However, their method works only for top-view images where the camera is perpendicular to the earth. This is a significant limitation for a surveillance scenario, where the camera view angle should be actively adjusted.

Although the mentioned methods above are able to find both point and contextual anomalies, they do not get a contextual input explicitly, except UAV-ADNet. Instead, they implicitly extract contextual information from a given input. Unlike others, CADNet is able to learn conditional normality during the training thanks to feeding the network with context attributes as additional inputs.


Despite the surveillance domain, there is a variety of works related to anomaly detection in visual data. The main approach in these methods is learning the normality for the image dataset during the training. Then, the trained model reconstructs a test image. The difference between the original test image and the reconstruction indicates anomalies in the image. There are different methods to learn the normality of data such as autoencoders and their variants (\cite{hasan2016learning, bozcanilkeruavadnet, zhou2017anomaly,  tuluptceva2020anomaly}), one-class SVMs (\cite{chalapathy2018anomaly, ruff2018deep}), and generative adversarial network (\cite{schlegl2017unsupervised, dong2020experimental}). These methods cannot incorporate the context for learning the normality. Moreover, they conduct anomaly detection at pixel-level. Therefore, they lack of explicit representation of which object instances are anomalous in images.

\subsection{Contributions}
The main contributions of our work can be summarized as:
\begin{itemize}
    \item \textbf{Conditional deep autoencoder to find contextual anomalies:} We propose a conditional variational autoencoder (CVAE)-based anomaly detector (CADNet) to detect the abnormal presence of objects in an environment. Unlike other studies, our method can predict contextual anomalies that are conditioned on the \IB{contextual attributes} (i.e., GPS and time). 
    \item \textbf{Analysis of contextual attributes to anomaly detection performance} We conduct experiments removing features from the network systematically in order to show how contextual attributes affect the performance. We show that the context-awareness is crucial to find anomalies. 
\end{itemize}{}


\begin{figure*}[!hbt] 
\centerline{
\includegraphics[width=0.99\textwidth]{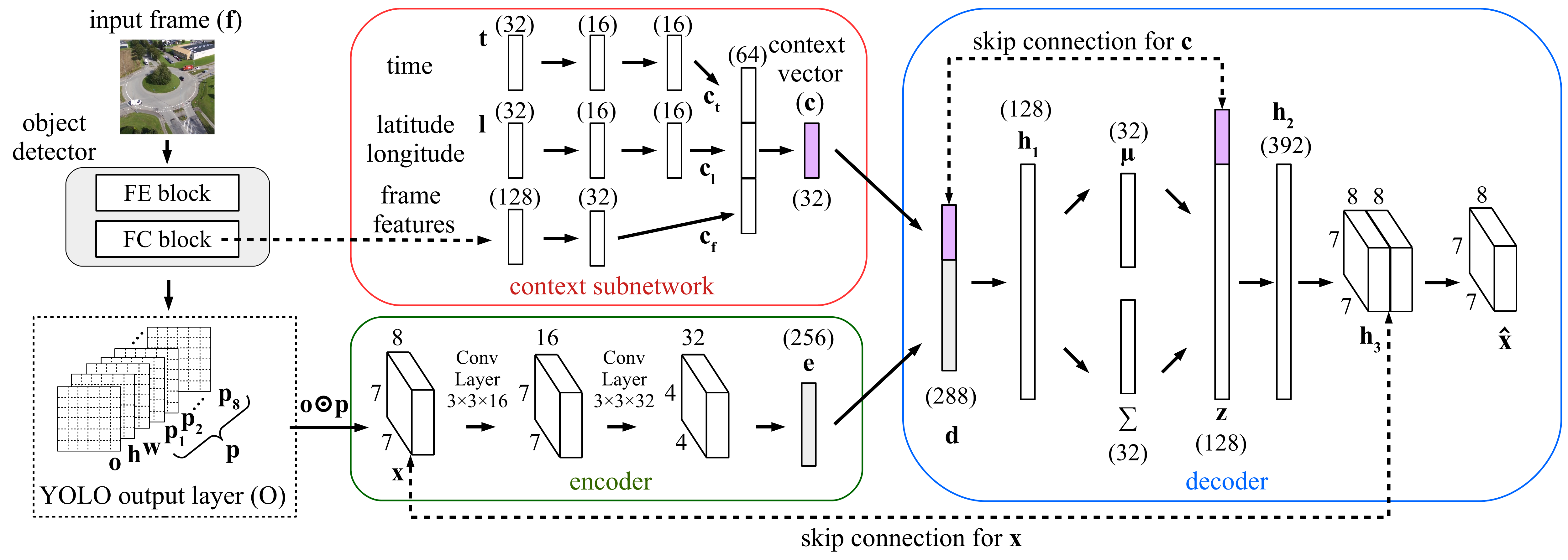}
}
\caption{This illustration shows the proposed architecture for contextual anomaly detection. The network consists of an encoder, a decoder, and a context subnetwork. The encoder takes a slice of YOLO-like object detector's output as input ($\mathbf{x}$), and the context subnetwork takes a time ($\mathbf{t}$) and GPS labels ($\mathbf{l}$) as input to produce a context vector ($\mathbf{c}$). Then, the decoder produces a reconstructed sample ($\mathbf{\hat{x}}$) for the given encoder's output conditioned on $\mathbf{c}$. [Best viewed in color]\label{fig:architecture}}
\end{figure*}

\section{\IB{CADNET}}

\subsection{\IB{Approach}}
\IB{Our study follows the main paradigm of anomaly detection: learning which patterns are normal in an environment rather than learning anomalies. The network is trained with normal samples (i.e., anomaly-free) collected from predefined monitoring-points. Therefore, it can learn what should be expected for a given monitoring point.}

\IB{Being a reconstruction-based method, the model uses the reconstruction error to detect anomalies in data. During the inference, it takes a data point, including anomalies, and tries to reconstruct it. For anomalies, the reconstruction error is higher than that of the normal data. Since it uses the reconstruction error as an anomaly score, a high reconstruction performance is necessary to prevent false predictions regarding what is normal or abnormal.}

\subsection{Network Architecture}
The network consists of three sub-networks: (i) encoder, (ii) decoder, and (iii) context subnetwork (Fig. \ref{fig:architecture}). The object detector predictions and the contextual attributes are streamed from the encoder and the contextual subnetwork, respectively. The outputs of these two sub-networks are concatenated to a vector and fed to the decoder. The decoder produces the reconstructed sample for given inputs.

As a preliminary step, any YOLO-like object detector can be used as an off-the-shelf backbone, which is not the focus of the research in this work. The object detector takes an image as input and produces a tensor including, bounding boxes and class predictions. The output feature map has a size of $S \times S \times (B * 5 \times C)$, where $S$ is the grid size (i.e., number of rows and columns of the feature map), $B$ is the number of bounding box predictions for each grid cell, and $C$ is the number of object classes. Each bounding box \IB{prediction} consists of 5 scalars, including the pixel coordinates represent the center of the box, that is relative to the grid cell location, the height and width of the box, and the confidence score that reflects the presence or absence of an object of any class. \IB{Non-maximum suppression} (NMS) is applied to the output feature map to prune imprecise bounding boxes.

The CADNet takes a tensor slice from an object detector's output as the main input. After NMS, the class probabilities are multiplied by the objectness score for each grid cell, and \IB{the} input tensor ($\mathbf{x}$) is created for CADNet. We discard bounding box offset scalars (e.g., the height and width of the bounding box) since we are only interested in the spatial layout of objects in an image frame rather than their sizes.

The encoder takes the input $\mathbf{x}$, and provide it into convolutional layers (Conv layers) to extract abstract features from the input. Then, the features are flattened to a 1D vector ($\mathbf{e}$) and fed into a fully connected layer (FC layer). 

The contextual attributes (time $\mathbf{t}$, latitude and longitude $\mathbf{l}$) are given to the context sub-network to extract contextual information of the given environment. Firstly, the time and GPS data are embedded into two vectors with a size of 32 using the FC layer. Then, two consecutive FC layers are applied to the embedded vectors to extract the time and GPS features, $\mathbf{c_t}$ and $\mathbf{c_l}$, respectively (see Fig. \ref{fig:architecture}). \IB{Besides the contextual attributes, the activations of the first FC layer of the object detector are also fed to FC layers in the context subnetwork to extract the frame features ($\mathbf{c_f}$). Then, they are concatenated with the context vector. In our experiments, we observe that this step is crucial for the network to make inferences for given input samples with different altitudes or camera angles but have the same timestamp or GPS coordinate.} 

The context vector ($\mathbf{c}$) and output of the encoder ($\mathbf{e}$) are concatenated into a vector ($\mathbf{d}$) and fed to the decoder. In the decoder, two groups of hidden activations (i.e., latent spaces) are calculated as in VAE, $\boldsymbol{\Sigma}$ and $\boldsymbol{\mu}$. The regularizer term is added to the loss function to constrain the network with that $\boldsymbol{\Sigma}$ and $\boldsymbol{\mu}$ correspond to variance and mean, respectively, of a prior distribution. For the sake of convenience, we assume a prior distribution as Gaussian distribution in our method. Furthermore, we sample hidden variables $\mathbf{z}$ using Gaussian distribution with the recently calculated parameters $\boldsymbol{\Sigma}$ and $\boldsymbol{\mu}$. The activation vector $\mathbf{z}$ is concatenated with the context vector $\mathbf{c}$ and fed into an FC layer to get features $\mathbf{h_2}$. Then, the vector $\mathbf{h_2}$ is reshaped into a tensor ($\mathbf{h_3}$) that has the same size as the original input $\mathbf{x}$ of the encoder. Then, the input $\mathbf{x}$ is copied and cropped to $\mathbf{h_2}$. As a final step, the convolutional filter with a size of $3 \times 3 \times 8$ is applied to the concatenated vector with the stride of 1. The output of the convolution is used as the reconstructed sample ($\mathbf{\hat{x}}$).

\IB{Note that the hidden vector $\mathbf{z}$ is sampled from the Gaussian distribution with the learnable parameters $\mathbf{\Sigma}$ and $\mathbf{\mu}$. The skip connections recover information lost during the sampling. Therefore, they help the recovery of the original encoder input.} In the network architecture, we employ skip-connections to propagate information from \IB{the} early layers to \IB{the} late layers to prevent information loss. To this end, we copy the input vector ($\mathbf{x}$) and crop it into the last activation map of the decoder. Then, a Conv Layer produces the reconstructed sample. Moreover, we also use skip connections for the context vector ($\mathbf{c}$), as in Fig. \ref{fig:architecture}. Therefore, the decoder produces a reconstructed sample $\mathbf{\hat{x}}$ for given input $\mathbf{x}$ conditioned on \IB{the} context vector $\mathbf{c}$. The reconstructed sample $\mathbf{\hat{x}}$ has the same structure as the input vector $\mathbf{x}$. The inference time (i.e., the forward pass starting from the encoder and the context subnetwork) of CADNet is calculated as 40 milliseconds on average on NVIDIA Jetson TX2.

\subsection{Loss Function}
Anomaly data instances are sufficiently far from normal data distribution. For contextual anomalies, boundaries of normal regions change according to the contextual attributes. To formulate these \IB{phenomena}, the network is subject to learn \IB{a} conditional normal data distribution. Therefore, during  training,  we  optimize  the  following loss function:
\begin{multline}
\mathcal{L}  = \frac{1}{N} \sum^N_{i=1} (\mathbf{x^{(i)}} \log{f(\mathbf{x^{(i)}}, \mathbf{t^{(i)}},\mathbf{l^{(i)}}) }\\ 
+(\mathbf{1}-\mathbf{x^{(i)}})\log{(\mathbf{1}-f(\mathbf{x^{(i)}}, \mathbf{t^{(i)}}, \mathbf{l^{(i)}}))}\\
+ {{D}_\text{KL}}[q(\boldsymbol{\mu}^{(i)}, \boldsymbol{\Sigma}^{(i)} \mid \mathbf{x^{(i)}}, \mathbf{c^{(i)}})\mid\mid p(\mathbf{z^{(i)}})]),
\end{multline}
where $\mathbf{x, t, l, c}$ denote the main input, time, GPS, and context vectors, respectively. $f$ indicates the model's output for given inputs, and $q$ denotes the probability distribution of the model parameters $\boldsymbol{\Sigma}, \boldsymbol{\mu}$, for given inputs, and it is modeled by the network. ${D}_\textbf{KL}$ is the Kullback Leibler (KL) divergence \cite{kullback1951information} that \IB{measures} the discrepancy between the $q$ distribution that is defined by the model's parameters and $p(\mathbf{z})$  that is the Gaussian distribution as the prior. Note that the loss function consists of the reconstruction loss (i.e., cross-entropy loss) and the regularizer term (i.e., KL loss).

\subsection{Training and Inference}
To train the network, the RMSProp optimizer \cite{tieleman2012lecture} is used with the initial learning rate \IB{0.05}. The learning rate is factorized by \IB{0.1} at epochs 5 and 18. The training is conducted on mini-batches with the size of 64 and ended the validation error starts to increase.

\IB{For inference, the network reconstructs the encoder input by a forward-pass. The cells corresponding to anomalies have lower values in the reconstructed sample, although they have a value of 1 in the original input. Element-wise difference between the original input tensor $x$ and the reconstructed tensor $\hat{x}$ indicates anomalies. If the reconstruction error for a cell is greater than a threshold, the corresponding object is considered an anomaly. We empirically set the decision threshold 0.6 in our experiments.}

\section{EXPERIMENTS AND RESULTS}
We conduct point and contextual anomaly detection tests on the AU-AIR traffic surveillance dataset \cite{bozcan2020air}. In our experiments, firstly, we compare the models' reconstruction performances since it is crucial for reconstruction-based anomaly detection tasks. Then, we compare CADNet with the state-of-the-art approaches for anomaly detection. Moreover, we conduct an ablation study to understand how contextual attributes affect the model's performance. 

\subsection{Dataset}
We train our model with the AU-AIR dataset \cite{bozcan2020air} that is an aerial surveillance dataset, including multi-modal flight data. The AU-AIR dataset consists of image frames along with bounding box annotations of traffic-related objects, and current GPS coordinate, time, altitude, and velocity of a UAV. The frames are captured in an outdoor environment for traffic surveillance (See Fig.  \ref{fig:intro}). Eight object categories (car, pedestrian, van, truck, bicycle, motorbike, trailer, bus) are annotated in the frames. The dataset includes 32,823 samples in total. During training, we split the dataset into three parts: \%60 for training, \%10 for validation, and \%30 for testing.


\subsection{Object Detection Model Settings}
Since object detection is not the main research focus in this paper, we use Tiny-YOLO V3 as the object detector. We choose Tiny-YOLO V3 since it requires less computation power and memory  \IB{making} it applicable for real-time inference on-board computers (i.e., 17 FPS on NVIDIA Jetson TX2. We use the weights of Tiny-YOLO V3 pretrained on the COCO dataset \cite{lin2014microsoft}, and train it using the AU-AIR dataset to find objects in aerial images.

\subsection{Baselines}

We compare CADNet with several approaches, including a standard autoencoder \cite{zhou2017anomaly}, one-class SVM \cite{chalapathy2018anomaly}, and GAN \cite{schlegl2017unsupervised}, to learn normality. For the sake of compatibility, we implement these baselines with the same architecture of the encoder-decoder parts of CADNet. Moreover, to compare the effect of context attributes and the network architecture to anomaly detection task, we systematically remove parts of the input to see which context attributes are relevant to the network's performance.





\subsection{Experiment 1: Model's Reconstruction Performances}
Since our anomaly detection method is based on a comparison of the original input and the model reconstruction, the network should be able to reconstruct the input data with a high reconstruction performance. The reconstruction is a challenging task when inputs are sparse matrices, as in our case.

We calculate an error on the difference between the original  data in the test set and the reconstructed counterpart, and observe during training phase:
\begin{equation}
    {E}_{train} = \frac{1}{S} \sum_{i=1}^{S}{\left({\sum_{m,n,l}^{M,N,L}\sqrt{\left({x_{m,n,l}^{(i)}} - {{\hat{x}_{m,n,l}}^{(i)}}\right)^2} }\right)},
\end{equation}
where S is the number of test samples in the dataset; $x_i$ and $\hat{x}_i$ are the original and reconstructed samples with the index $i$, respectively; $M, N, L$ are the number of rows, columns, and depth in $x$ and $\hat{x}$, respectively. 

Table \ref{tbl:results} compares the reconstruction performances of CADNet and the baseline approaches. As can be seen in the table, CADNet has lower reconstruction error than others. This finding implies that CADNet is more promising for the anomaly detection task compared to other baselines since having lower reconstruction performance is crucial for the task.



\subsection{Experiment 2: Point Anomaly Detection}
Initially, our model is tested for point anomalies. In the context of surveillance, point anomalies can be considered as observations that break predefined rules for an environment. These rules are valid under any circumstances (e.g., a car cannot drive on bike road, a pedestrian cannot walk on a motorway). We create a set of traffic rules (12 rules in total, 8 for vehicles and 4 for pedestrians) according to the current traffic regulations in Denmark \cite{undervisning}. 

\begin{figure}[!t] 
\centerline{
\includegraphics[width=0.44\textwidth]{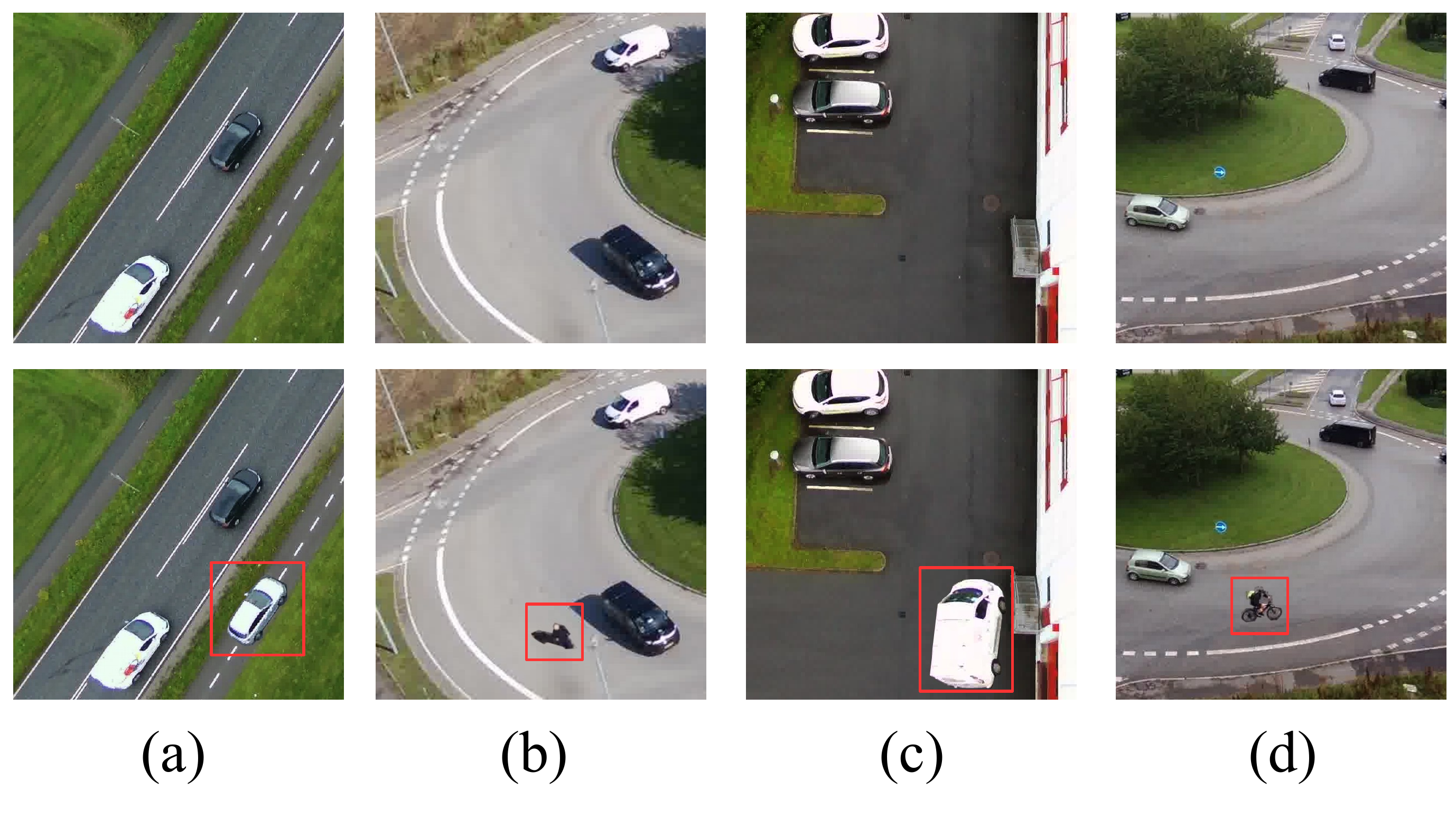}
}
\caption{Examples of created anomalies in the test dataset: (a) a car on a bike road, (b) a person on a road, and contextual anomalies such as (c) a parked van in front of a building, (d) a bicycle on a road. \label{fig:anomaly_pathces}}
\end{figure}

Ground-truth labels for anomalies are not available in the dataset, like in most of the real-world scenarios. Therefore, we add objects manually to the samples to create an evaluation set, including point anomalies. To this end, we extract patches of object instances from images manually and paste them to the test images. We apply the object detector to the generated samples to observe whether it can find newly pasted objects or not. We discard the image samples in which the object detector cannot find added objects. As a result, we create 180 test samples with the point anomalies in total. Examples from generated samples can be seen in Fig. \ref{fig:anomaly_pathces}. \IB{We define accuracy as the percentage of anomalies correctly estimated with respect to the ground-truths in the test dataset for this task.}

Table \ref{tbl:results} provides point anomaly detection results of CADNet and the baselines. As shown in the table, CADNet is better than other approaches in finding point anomalies. Surprisingly, the autoencoder and GAN have lower reconstruction errors than one-class SVM, yet their point anomaly detection results are significantly lower. We observe that the autoencoder and GAN tend to learn identity function. Therefore, they cannot entirely remove anomalies when they are presented in test data.

Moreover, we evaluate this task with CADNet and its variants (See Table \ref{tbl:results}). We see that CADNet has the highest accuracy. However, there is no significant difference in the case that contextual attributes are not fed to the networks (CADNet-wo-gps-time). Naturally, point anomalies do not depend on the contextual attributes, and they are considered as anomalies regardless of current time and location. Furthermore, we observe that skip-connections are also crucial for anomaly detection performance. The model without skip connections (CADNet-wo-skip) has lower accuracy compared to the other models. Similar to the phenomenon in the reconstruction test, the skip connection improves performance preventing loss of information through layers.



\begin{table}[!t]
\footnotesize 
\caption{Comparison with baselines. 
\label{tbl:results}}
\begin{center}
\begin{tabular}{c|ccc}\hline
Model & Recon.  &Point  & Contextual  \\ 

Name &  error &  anomaly acc. &  anomaly acc.  \\
\hline \hline

CADNet & $0.00$ & $91.2\%$& $86.6\%$\\
Autoencoder & $0.02$ & $17.1\%$& $21.4\%$\\
One-class SVM &  $0.09$ & $72.3\%$& $69.2\%$\\
GAN &  $0.03$ & $48.3\%$& $51.2\%$\\
\hline \hline
\multicolumn{4}{c}{\textit{Ablation experiments follow below}} \\
\hline \hline
CADNet-wo-gps-time &  $0.00$ & $90.2\%$& \IB{$37.2\%$}\\
CADNet-wo-time & $0.00$ & $91.6\%$ & $54.4\%$\\
CADNet-wo-gps &$0.00$ & $88.7\%$ & $63.8\%$\\
CADNet-wo-frame &$0.00$ & $88.0\%$ & $78.6\%$\\
CADNet-wo-skip & $0.17$ & $37.7\%$& $35.0\%$\\
CADNet-wo-skip-m &$0.17$ & $42.3\%$ & $32.9\%$\\
CADNet-wo-skip-c &  $0.00$ & $88.5\%$& $76.9\%$ \\ \hline
\end{tabular}
\end{center}

\end{table}

\subsection{Experiment 3: Contextual Anomaly Detection}
We also test our model with contextual anomalies that are more challenging than point anomalies. In the context of surveillance, the contextual anomalies do not necessarily break any rules. Nonetheless, they arise suspicions due to being less likely to occur for particular environmental settings. The nature of contextual anomalies makes test data generation not straightforward compared to point anomalies. Therefore, we create a test set in a different way than point anomalies.

We separate the test frames into the groups according to the GPS labels and the altitudes of their monitoring points, and the monitoring time. Examining the presence and layouts of objects in each group, we add anomolous objects, which should not appear in the images for a specific observation time or a monitoring point. The added objects do not violates any predefined rules for an environment. However, they still should be considered anomaly since they are different than normal observations. We have 120 test samples in total including contextual anomalies.

As can be seen in Table \ref{tbl:results}, CADNet is superior to find contextual anomalies in an environment. This is expected since feeding CADNet with contextual attributes allows it to find the anomalies conditioned on the context. Similar to the results in point anomaly detection, the autoencoder and GAN have lower accuracy than the one-class SVM.


\subsection{Experiment 4: Effect of Contextual Attributes to Anomaly Detection}
We systematically remove parts from the original architecture of CADNet and train the models to analyze the effect of contextual attributes and frame input to the anomaly detection performance in our surveillance case. \IB{Firstly, we create CADNet-wo-gps-time model removing both the GPS coordinates and time inputs from the model. We observe that the contextual anomaly detection performance decreases dramatically (from 86.6\% to 37.2\%); although the performance change in the point anomaly detection task is not significant (from 91.2\% to 90.2\%). This result supports that the contextual attributes are essential for the network to make a conditional inference. We also create CADNet-wo-time and CADNet-wo-gps models, removing the GPS and time inputs from the model, respectively, to observe the effect of contextual attributes separately. Although there is no notable change for the point anomaly detection task, we see that the contextual anomaly detection performance is lower when the time input is removed since contextual anomalies depend on GPS more than on time. }

\IB{Next, we remove the frame activation input (CADNet-wo-frame) to observe its effect to the anomaly detection tasks. Although the performance change in the point anomaly detection task is not notable, the results in contextual anomaly detection task is interesting. CADNet-wo-frame has a higher accuracy (\%78.6) than any model without GPS or time inputs. We can interpret this observation as that the frame activations are not as effective as the GPS and time inputs to the contextual anomaly detection performance. However, the lack of the frame activations still reduces the contextual anomaly detection performance compared to the original CADNet. We observe that the network's contextual anomaly detection performance is lower for test samples collected from monitoring points which have same GPS labels but the drone rotates in the air, or the drone has different altitudes. These finding suggests that the frame activations can be considered an contextual attribute which gives information regarding the visual appearance of a scene.}

\IB{We analyze the effect of the skip connections to the anomaly detection performance. When we remove the skip connections both for the main input (i.e., the output layer of the object detection) and the context vector (CADNet-wo-skip), the performance for point and contextual anomaly defections drop significantly (37.7\% and 35.0\%, respectively). Moreover, the network's reconstruction error increases by 0.17. This fact is a result of the lack of recall capability without a skip connection since the sampling step causes an information loss, and the skip connections carry information from early layers to late layers skipping the sampling step.} 

\IB{We also analyze the effects of the skip connections for the main input and the context vector separately. Firstly, we discard the skip connection for the main input, namely CADNet-wo-skip-m. The point and contextual anomaly detection results are significantly lower than CADNet. Then, we remove the skip connection for context vector (CADNet-wo-skip-c). We observe that CADNet-wo-skip-c has lower performance than CADNet as expected. Moreover, the model's performance on the contextual anomaly detection task drops significantly (from 86.6\% to 76.9\%) compared to the point anomaly detection task (from 91.2\% to 88.5\%). This is expected since the presence of the skip connection for the contextual attributes is more crucial in contextual anomaly detection task, where the network makes inferences conditioned on the contextual attributes.}

\section{DISCUSSION}
\IB{The training and test samples are collected from predefined monitoring-points over the interest area. Therefore, background objects (e.g., roads) will have the same layout as long as images are recorded at the same monitoring-point. The network then learns the layout of foreground objects (e.g., vehicles, pedestrians) in the image plane. Since the encoder input does not have any visual clue on the image appearance, background changes in images cannot affect the encoder input. }

\IB{The form of the encoder input might cause a practical limitation for aerial surveillance scenarios in which the drone should monitor previously unseen regions of an environment. An unseen region might have novel object layouts. Therefore, CADNet might be expected to fail in detecting anomalies in this region since it is not trained to learn what the normality is for that area. This problem can be solved at the application level by collecting training data from new predefined way-points to increase surveillance coverage.}

\begin{figure} [ht]
    \centering
  \subfloat[\IB{False positive}\label{fig:results-sim}]{%
        \includegraphics[width=0.23\textwidth]{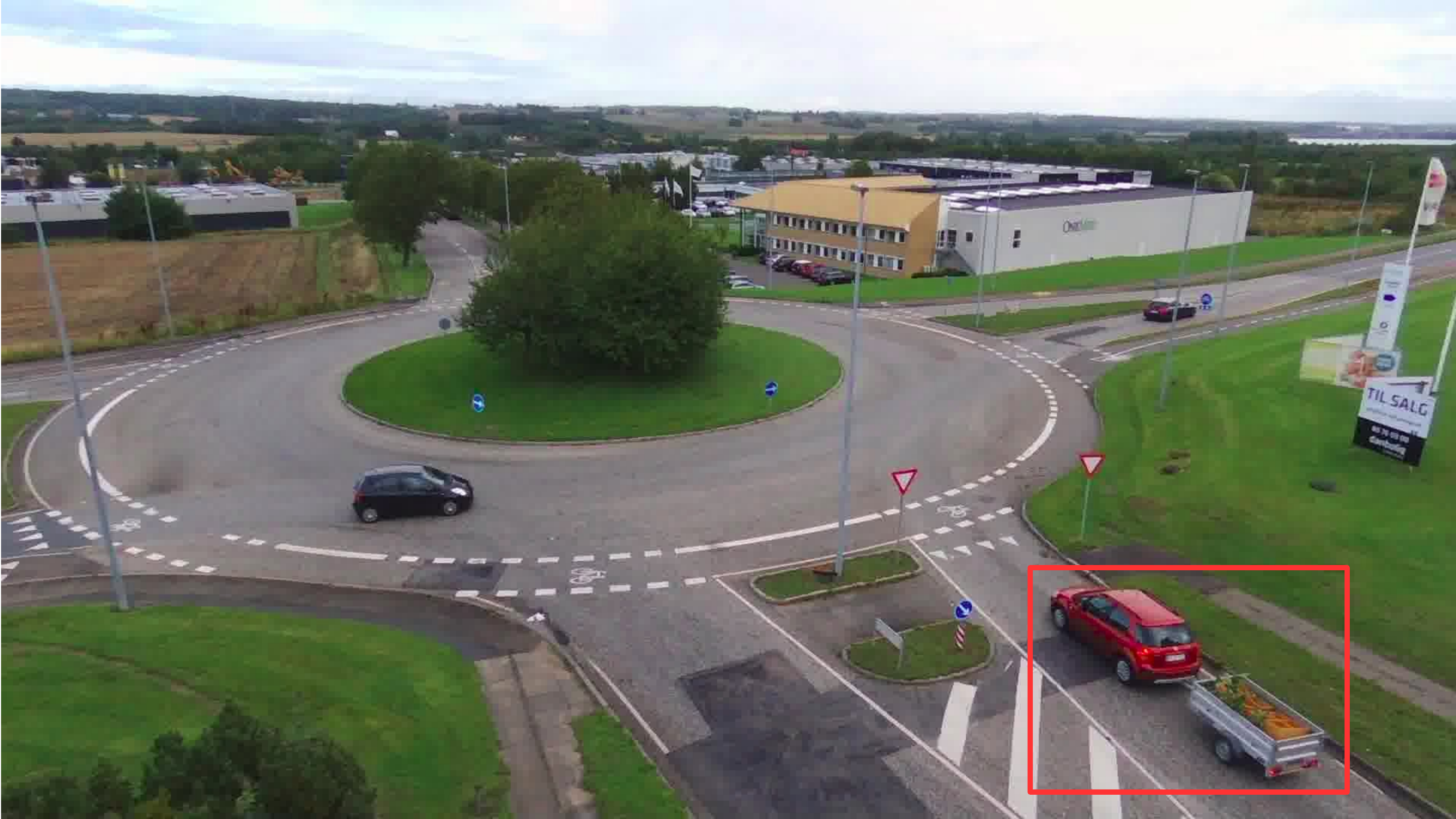}}\hspace{4pt}
  \subfloat[\IB{False negative}\label{fig:results-real}]{%
       \includegraphics[width=0.23\textwidth]{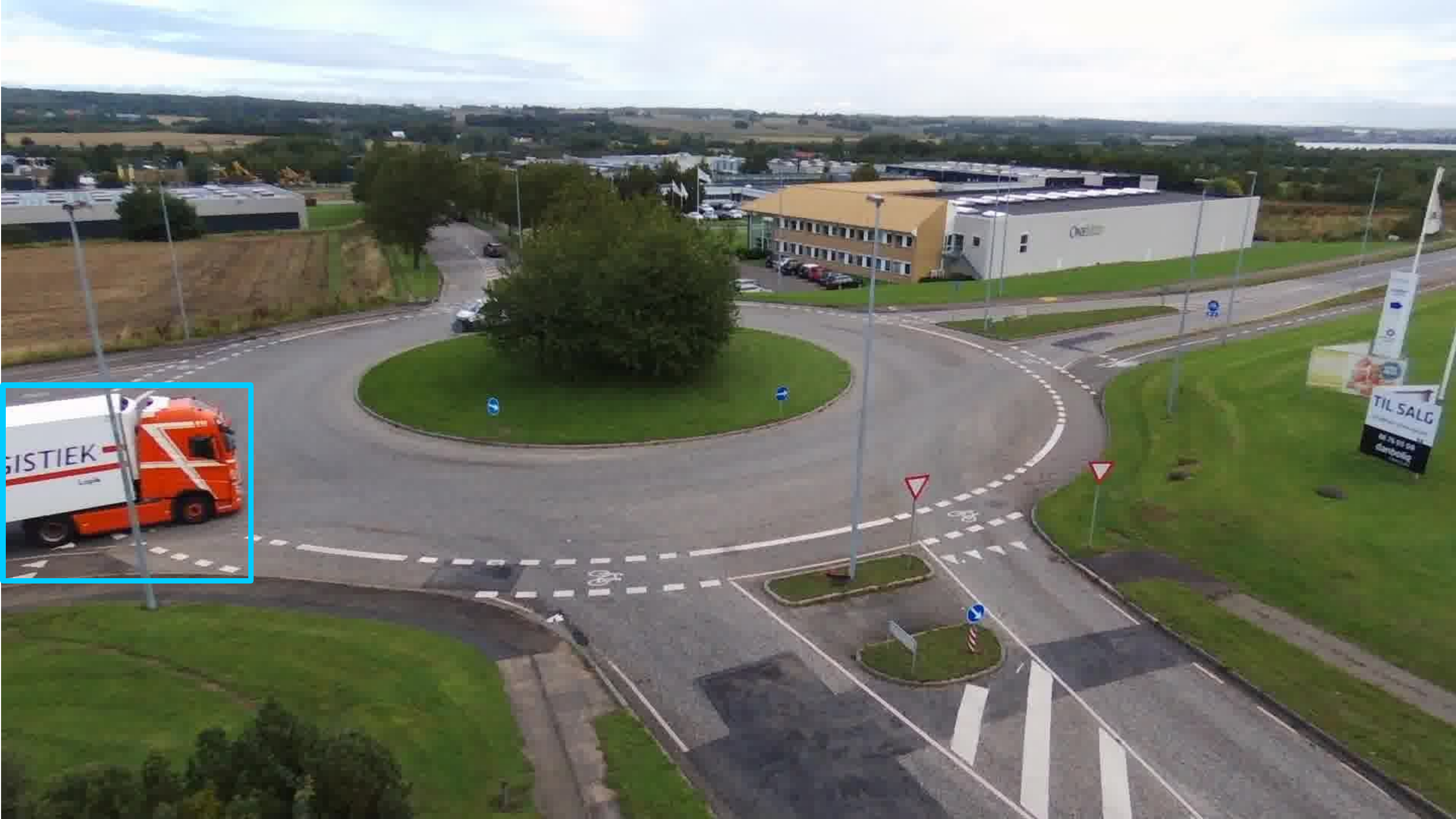}}
  \caption{\IB{Failures of the proposed approach. (a) The object detector detects a car with a trailer as a truck. Therefore, CADNet recognizes it as an anomaly by mistake. In (b), CADNet cannot predict a truck as an anomaly since it is detected as a van by the object detector.}}
  \label{fig:false_predictions} 
\end{figure}

\IB{CADNet works on top of YOLO-like object detectors. Therefore, it can be affected by false predictions of object detectors. In case of an object detection failure, CADNet might miss objects that should be detected as an anomaly, or it can falsely consider an object as abnormal (See Fig. \ref{fig:false_predictions}).}

\section{CONCLUSIONS}
We propose a contextual anomaly detection method (CADNet) based on a deep neural network. Unlike similar studies, our network works on the top of object detectors and finds object-wise anomalies in the environment. In our experiments, we compare our method with existing approaches to find point and contextual anomalies for low altitude aerial surveillance scenerio. As a result, we show our method's superiority for the anomaly detection tasks. The inference time is 40 milliseconds on average on NVIDIA Jetson TX2, which is applicable for real-time applications.

We conduct an ablation study to see how contextual attributes and the network's components affect the model's performance. In our experiments, we observe that the model can learn anomalies conditioned on contextual attributes, and show that the contextual attributes are crucial for finding context-dependent anomalies.

\bibliographystyle{IEEEtran}

\bibliography{references}

\end{document}